\documentclass[sigconf,screen]{acmart}
\AtBeginDocument{%
 }

\copyrightyear{2023}
\acmYear{2023}
\setcopyright{acmlicensed}\acmConference[MM '23]{Proceedings of the 31st
ACM International Conference on Multimedia}{October 29-November 3,
2023}{Ottawa, ON, Canada}
\acmBooktitle{Proceedings of the 31st ACM International Conference on
Multimedia (MM '23), October 29-November 3, 2023, Ottawa, ON, Canada}
\acmPrice{15.00}
\acmDOI{10.1145/3581783.3612366}
\acmISBN{979-8-4007-0108-5/23/10}

\begin{document}

\title{KeyPosS: Plug-and-Play Facial Landmark Detection through GPS-Inspired True-Range Multilateration}


\author{Xu Bao}
\authornote{Denotes equal contributions}
\email{baoxu@email.szu.edu.cn}
\affiliation{%
  \institution{DAMO Academy, Alibaba Group}
  \country{}
}

\author{Zhi-Qi Cheng}
\authornotemark[1]
\authornote{Zhi-Qi Cheng and Jun-Yan He are the corresponding authors}
\email{zhiqic@cs.cmu.edu}
\affiliation{%
  \institution{Carnegie Mellon University}
  \country{}
}

\author{Jun-Yan He}
\authornotemark[1]
\authornotemark[2]
\email{leyuan.hjy@alibaba-inc.com}
\affiliation{%
  \institution{DAMO Academy, Alibaba Group}
  \country{}
}

\author{Wangmeng Xiang}
\email{wangmeng.xwm@alibaba-inc.com}
\affiliation{%
  \institution{DAMO Academy, Alibaba Group}
  \country{}
}

\author{Chenyang Li}
\email{lichenyang.scut@foxmail.com}
\affiliation{%
  \institution{DAMO Academy, Alibaba Group}
  \country{}
}

\author{Jingdong Sun}
\email{jingdons@andrew.cmu.edu}
\affiliation{%
  \institution{Carnegie Mellon University}
  \country{}
}

\author{Hanbing Liu}
\email{liuhb21@mails.tsinghua.edu.cn}
\affiliation{%
  \institution{Tsinghua University}
  \country{}
}

\author{Wei Liu}
\email{ustclwwx@gmail.com}
\affiliation{%
  \institution{DAMO Academy, Alibaba Group}
  \country{}
}

\author{Bin Luo}
\email{luwu.lb@alibaba-inc.com}
\affiliation{%
  \institution{DAMO Academy, Alibaba Group}
  \country{}
}

\author{Yifeng Geng}
\email{cangyu.gyf@alibaba-inc.com}
\affiliation{%
  \institution{DAMO Academy, Alibaba Group}
  \country{}
}

\author{Xuansong Xie}
\email{xingtong.xxs@taobao.com}
\affiliation{%
  \institution{DAMO Academy, Alibaba Group}
  \country{}
}

\renewcommand{\shortauthors}{Xu Bao et al.}

\begin{abstract}
Accurate facial landmark detection is critical for facial analysis tasks, yet prevailing heatmap and coordinate regression methods grapple with prohibitive computational costs and quantization errors. Through comprehensive theoretical analysis and experimentation, we identify and elucidate the limitations of existing techniques. To overcome these challenges, we pioneer the application of True-Range Multilateration, originally devised for GPS localization, to facial landmark detection. We propose KeyPoint Positioning System (KeyPosS) - the first framework to deduce exact landmark coordinates by triangulating distances between points of interest and anchor points predicted by a fully convolutional network. A key advantage of KeyPosS is its plug-and-play nature, enabling flexible integration into diverse decoding pipelines. Extensive experiments on four datasets demonstrate state-of-the-art performance, with KeyPosS outperforming existing methods in low-resolution settings despite minimal computational overhead. By spearheading the integration of Multilateration with facial analysis, KeyPosS marks a paradigm shift in facial landmark detection. The code is available at \hyperlink{blue}{https://github.com/zhiqic/KeyPosS}.
\end{abstract}

\ccsdesc[500]{Computing methodologies~Facial landmark detection}
\keywords{Facial Landmark Detection, True-Range Multilateration}

\maketitle

\begin{figure} [!ht]
\small
	\begin{center}
	\includegraphics[width=0.65\linewidth]{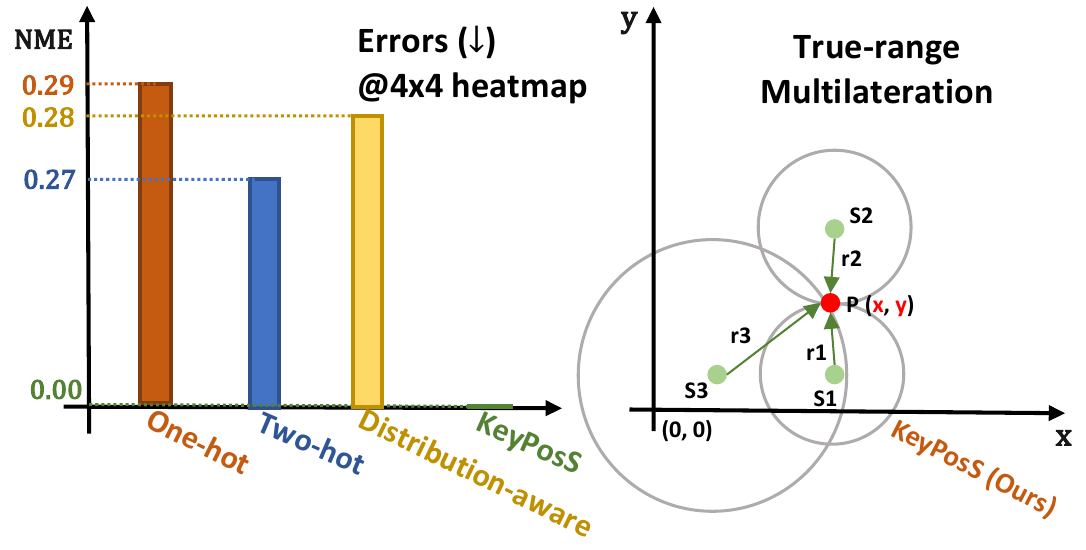}
	\end{center}
	\vspace{-0.15in}
 	\caption{\small
Comparison of the facial landmark detection strategies. On the left, we highlight the accuracy of our method, KeyPosS, which employs True-Range Multilateration, surpassing the previous one-hot and two-hot decoding approaches. The right side visualizes the main idea of the True-Range Multilateration algorithm, initially used in GPS systems and now adapted for facial
landmark detection.}
  \vspace{-0.1in}
        \label{fig:compare}
	\vspace{-0.1in}
\end{figure}

\begin{figure*} [!ht]
\small
	\begin{center}
		\includegraphics[width=0.7\linewidth]{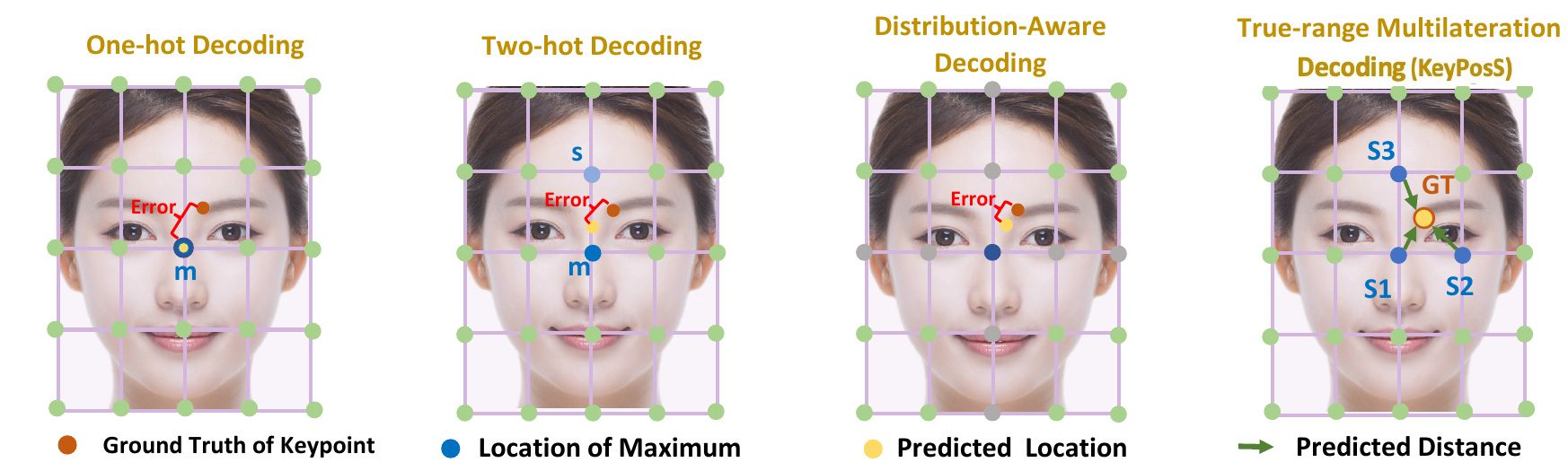}
	\end{center}
	\vspace{-0.15in}
	\caption{\small  Comparison of four decoding methods: basic one-hot, two-hot, distribution-aware, and our proposed KeyPosS. Despite the inherent "Error" in encoding-decoding, KeyPosS excels with minimal overhead and almost no added computational load.}
	\label{fig:encoding}
	\vspace{-0.1in}
\end{figure*}

\section{Introduction}
Facial landmark detection is fundamental for enriching personalized e-commerce interactions~\cite{cheng2016video, cheng2017video, cheng2017video2shop, cheng2017selection, sun2018personalized, nguyen2017vireo}, providing insights into nuanced human analysis~\cite{he2021db, tu2023implicit,zhou2022hypergraph, zhou2023overcoming, chen2023hdformer}, and enhancing the precision of various biometric recognition techniques~\cite{huang2018gnas, cheng2018learning, cheng2019improving, cheng2019learning, huang2020stacked, cheng2022gsrformer, cheng2022rethinking, lan2023procontext,li2023longshortnet, he2023damo,liu2023posynda,liu2023refined}.

Despite some progress~\cite{Newell_Hourglass_ECCV16, Nibali_Corr_18, Cao_TPAMI_21}, the previous methods heavily rely on heatmap~\cite{Newell_Hourglass_ECCV16,Chandran_attention_CVPR2020,Ronneberger_UNet_MICCAI15,Sun_HRNet_CVPR19,Tang_Quan_ECCV18,Zhang_darkpose_CVPR20,Zou_Robust_ICCV19,Wang_AWingLoss_ICCV19,Jin_PIPNet_IJCV21} or coordinate regression~\cite{Dapogny_DeCaFA_ICCV19,Zhu_Coarse2Fine_CVPR15,Lv_TSR_CVPR17,Feng_Wingloss_CVPR18,Lv_Regression_CVPR17,Li_Structure_ECCV20,Wu_LaB_CVPR18,Nibali_Corr_18} techniques, leading to challenges such as computational burden and quantization errors.
Typically, as shown in Figure~\ref{fig:compare}-\ref{fig:encoding}, the heatmap-based methods often have quantization errors when heatmaps are downscaled from their original input~\cite{Newell_Hourglass_ECCV16}, while large-resolution heatmaps become computationally expensive~\cite{Ronneberger_UNet_MICCAI15}. Conversely, coordinate regression-based methods attempt to reduce the computational complexity, but always lack the necessary spatial and contextual information, complicating the task due to inherent visual ambiguity~\cite{Feng_Wingloss_CVPR18}.

To address these issues, we present the KeyPoint Positioning System (KeyPosS) — a groundbreaking facial landmark detection paradigm depicted in Figure~\ref{fig:pipeline}. Uniquely, KeyPosS employs the True-range Multilateration algorithm, initially used in GPS systems, for rapid and accurate facial landmark detection. This framework utilizes a fully convolutional network to predict a distance map, which estimates the distance between a Point of Interest (POI) and multiple anchor points. These anchor points are ingeniously leveraged to triangulate the POI's position through the True-range Multilateration algorithm. Beyond its strategy, KeyPosS's plug-and-play nature ensures easy integration into existing models without the need for additional training or modifications. Its usability is further enhanced by its robust location accuracy, even with low-resolution distance maps, making it a versatile and adaptable solution for the challenges in facial landmark detection. To sum up, the invention of KeyPosS revolves around five key aspects:

\begin{enumerate}
    \item \textit{Optimizing heatmap potential:} KeyPosS utilizes the True-Range Multilateration algorithm to analyze both the main response and its neighboring anchor points on the heatmap. This dual principle improves accuracy in intricate scenarios like occlusions or limited visibility.
    
    \item \textit{Fusing regression and classification:} KeyPosS classifies POIs and anchors by regressing Gaussian-encoded distances. These dual-purpose tasks effectively represent distances and key points (POIs), thereby streamlining allocation.
    
    \item \textit{Robust localization via Station Anchor Sampling:} KeyPosS utilizes regional maxima and neighboring points, enhancing detection robustness and countering errors from isolated peak values.
    
    \item \textit{Plug-in Manner for heatmap methods:} The KeyPosS decoding scheme can conveniently retrofit existing heatmap-based methods, promoting adaptability and compatibility across various systems.
    
    \item \textit{Improved performance across resolutions:} KeyPosS exhibits better accuracy across varying distance map resolutions, underscoring its flexibility for diverse real-world contexts.
\end{enumerate}

In essence, KeyPosS sets a new benchmark for 2D landmark detection, surpassing contemporary techniques in versatility, scalability, and performance. This enhancement heralds its potential applicability in myriad 2D keypoint use cases, like sports analytics, strengthening their real-time efficiency and robustness.\footnote{The source code is available at https://github.com/zhiqic/KeyPosS}

\section{Related Work}
Facial landmark detection has been broadly studied~\cite{Milborrow_ASM_ECCV08,Hebert_AAM_TPAMI95,Cristinacce_CLM_PR08,Cao_ESR_IJCV14,Zhu_shape_CVPR15}. Particularly, current researches have shifted towards deep learning-based methods, which can be briefly classified into heatmap and coordinate strategies.

\noindent \textbf{Heatmap-based methods.}~These methods achieve accurate localization and impressive performance by leveraging high-resolution feature maps. The representative architectures like Stacked Hourglass Network~\cite{Newell_Hourglass_ECCV16} and UNet~\cite{Ronneberger_UNet_MICCAI15} have excelled in this domain~\cite{Newell_Hourglass_ECCV16,Chandran_attention_CVPR2020,Ronneberger_UNet_MICCAI15,Sun_HRNet_CVPR19,Tang_Quan_ECCV18,Zhang_darkpose_CVPR20,Zou_Robust_ICCV19,Wang_AWingLoss_ICCV19,Jin_PIPNet_IJCV21}.
HRNet~\cite{Sun_HRNet_CVPR19}, for instance, obtains high-resolution maps through multi-scale image feature fusion, demonstrating promising results. HSLE~\cite{Zou_Robust_ICCV19} proposes a hierarchical structure for accurate alignment, while previous work~\cite{Newell_Hourglass_ECCV16} compensates for quantization error. Darkpose~\cite{Zhang_darkpose_CVPR20} contemplates the heatmap distribution for more precise target locations, and PIPNet~\cite{Jin_PIPNet_IJCV21} performs simultaneous predictions, reducing latency. However, these methods' heavy reliance on high-resolution feature maps inevitably increases computational costs.

\noindent \textbf{Coordinate regression methods.} These approaches directly map discriminative features to target landmark coordinates~\cite{Dapogny_DeCaFA_ICCV19,Zhu_Coarse2Fine_CVPR15,Lv_TSR_CVPR17,Feng_Wingloss_CVPR18,Lv_Regression_CVPR17,Li_Structure_ECCV20,Wu_LaB_CVPR18,Nibali_Corr_18}. For example, DeCaFA~\cite{Dapogny_DeCaFA_ICCV19} integrates U-Net~\cite{Ronneberger_UNet_MICCAI15} to preserve spatial resolution in cascaded regression for face alignment. Lv et al.~\cite{Lv_Regression_CVPR17} introduce Two-Stage Reinitialization (TSR), a deep regression architecture that addresses initialization with a two-stage process, involving global and local stages for coarse and fine alignment. LAB~\cite{Wu_LaB_CVPR18} combines heatmap regression with coordinate prediction, employing facial boundaries as geometric constraints. A cascaded Graph Convolutional Network~\cite{Li_Structure_ECCV20} leverages both global and local features for accuracy. Additionally, Nibali et al.~\cite{Nibali_Corr_18} propose a differentiable spatial to numerical transform (DSNT) to address differentiation challenges in regression-based methods, ensuring spatial generalization.

\noindent\textbf{Comparison with previous works}.~KeyPosS is a one-of-a-kind system that uniquely leverages GPS-inspired true-range multi-point positioning technology for facial point-of-interest (POI) localization.
Unlike prior methods reliant solely on coordinate regression or heatmap-based approaches, KeyPosS amalgamates both regression and classification pipelines, providing a comprehensive solution for facial landmark detection. 
The plug-and-play nature of KeyPosS enables seamless integration into existing frameworks without requiring customization. 
The adaptability of KeyPosS indicates potential utility not only in facial landmark detection but also in a range of other keypoint detection tasks, suggesting a promising direction for future research.

\section{Methodology}
To better understand our proposed KeyPosS framework, we first review the encoding and decoding methods for facial landmark detection (Sec.~\ref{sec:encoding_decoding}). Following in Sec.~\ref{sec:keypoint_representation}, we first theoretically explore the impact of different keypoint representations, analyzing the quantization error causes of various encoding-decoding schemes. In Sec.~\ref{sec:error_analysis}, we also experimentally analyze the error upper bounds when different heatmap resolutions are used as the final output. Finally, Sec.~\ref{sec:keyposs_framework} presents a comprehensive description of our proposed KeyPosS framework.

\subsection{Review of Facial Landmark Detection}
\label{sec:encoding_decoding}
Facial landmark detection entails locating predefined landmarks on a human face within an image $I\in R^{H\times W\times 3}$. The heatmap-based detection pipeline can be defined as:
\begin{equation}
\mathcal{H} = \mathcal{F}(I,W)
\end{equation}
where $\mathcal{F}$ is the ConvNet backbone, $\mathcal{H}$ is the predicted heatmap, and $W$ is the backbone weight. $\mathcal{H}\in R^{h\times w\times K}$, with $h = H/\lambda$, $w=W/\lambda$, and $\lambda$ as the downsampling ratio. $K$ denotes the number of predefined facial landmarks. Generally, joint coordinate prediction involves finding the maximal activation location, but since the heatmap usually doesn't share the original image's spatial size, the heatmaps must be up-sampled using a sample-specific factor $\lambda$. This gives rise to a sub-pixel localization problem, typically solved through encoding-decoding pipelines. Gaussian encoding is a common encoding scheme encoding the distance from the point of interest (POI) to the station anchor. We will discuss the previous encoder and decoder methods in separate subsections.

\subsubsection{\textbf{Previous Encoding Methods}}
Traditional encoding strategies can be divided into the following two categories:

\noindent \textit{(1)~Unbiased Coordinate Encoding.}~To address noise from distant station anchors, unbiased coordinate encoding is widely used in previous works. After transforming the ground-truth distances between the station anchors and the POIs, the resolution reduction is defined as:
\begin{equation}
{\boldsymbol{\beta}} = \frac{{\boldsymbol{\beta}}}{\lambda} = \left(\frac{u}{\lambda}, \frac{v}{\lambda} \right).
\end{equation}
Next a spatially variant Gaussian function $\mathcal{G}(\cdot)$ and parameter $\sigma$ are used to encode the distance. Given a station anchor at $(x,y)$ in the heatmap, the distance encoding $\mathcal{G}(x,y, {\beta})$ is defined as:
\begin{equation}
\label{equ:gaussian}
\mathcal{G}(x,y, {\boldsymbol{\beta}}) = \frac{1}{2 \pi r}\exp{\left(-\frac{(x-u)^2 + (y-v)^2}{2 \sigma^{2}}\right)}.
\end{equation}
The motivation behind this is that this method decreases the distance map's sensitivity to station anchors that are far from the POI, providing more accurate distance predictions.

\noindent \textit{(2)~Biased Coordinate Encoding.}~Different from the previous unbiased coordinate encoding, biased coordinate encoding simply considers the heatmap ground truths as a category label, applying quantization during label assignment as:
\begin{equation}
{\boldsymbol{\beta}} = \text{Quan}(\frac{{\boldsymbol{\beta}}}{\lambda}) = \text{Quan}(\left(\frac{u}{\lambda }, \frac{v}{\lambda} \right))
\end{equation}
where $\text{Quan}$ signifies the quantization operation, and the representation is also encoded by Equation~\ref{equ:gaussian}.

\subsubsection{\textbf{Previous Decoding Methods}}
Common decoding schemes such as One-hot, Two-hot, and Compensation, are outlined here based on the predicted heatmap $\mathcal{H}$:

\noindent \textit{(1)~One-hot Decoding.} This strategy identifies the maximum response in $\mathcal{H}$:
\begin{equation}
\textbf{m} = \text{argmax}(\mathcal{H}),
\end{equation}
where $\textbf{m}$ is the location of the maximum value. The predicted keypoint $\hat{\boldsymbol{\beta}} = \boldsymbol{m}$.

\noindent \textit{(2)~Two-hot Decoding.} An enhanced version of One-hot decoding is Two-hot encoding \cite{Newell_Hourglass_ECCV16}, which uses both the first and second maximum value of heatmap $\mathcal{H}$:
\begin{equation} \label{equ:two-hot}
\hat{\boldsymbol{\beta}} = \textbf{m} + 0.25\frac{\textbf{m}-\textbf{s}}{||\textbf{m}-\textbf{s}||_{2}},
\end{equation}
where $\hat{\boldsymbol{\beta}}$ is the Two-hot decoding result, and $\textbf{s}$ represents the second maximum value's location. $||\cdot||_{2}$ is the vector magnitude.

\noindent \textit{(3)~Distribution-aware Decoding.} While the two-hot decoding method accounts for the second maximum response, the target position can't be precisely identified with just two points. The Darkpose approach \cite{Zhang_darkpose_CVPR20} examines the predicted heatmap's distribution structure to infer the underlying precision location:
\begin{equation}
\mathcal{G}(\boldsymbol{\beta}, \textbf{x} , \Sigma) = \frac{1}{2 \pi r}\exp{\left(-\frac{1}{2}(x-\boldsymbol{\beta})^{T}\Sigma^{-1} (x-\boldsymbol{\beta})\right)},
\end{equation}
where $\mathbf{x}$ is a pixel location in the predicted heatmap, $\hat{\boldsymbol{\beta}}$ is the Gaussian mean (center) corresponding to the estimated joint location, and the covariance $\Sigma$ is a diagonal matrix. The objective is to estimate the location $\hat{\boldsymbol{\beta}}$. The precision location can be approximated by evaluating a Taylor series (up to the quadratic term) at the maximal activation $m$ of the predicted heatmap. The details can be found in \cite{Zhang_darkpose_CVPR20}. The final equation is as follows:
\begin{equation}
\hat{\boldsymbol{\beta}} = D^{'}(m) + D^{''}(m),
\end{equation}
where $D^{'}$ and $D^{''}$ represent the first and second-order derivatives of the predicted heatmap, respectively.

\begin{table}
    \small
    \centering
    \caption{\small Experimental quantization error at different scales of the heatmap. The evaluation metric is Normalized Mean Error (NME, the smaller the better).}
    \vspace{-0.1in}
    \setlength{\tabcolsep}{1.35mm}{
    \begin{tabular}{l|c|c|c|c|c}
    \hline
    Resolution            & $64^2$  & $32^2$ & $16^2$ & $8^2$   & $4^2$  \\ \hline
    One-hot       & 0.018 & 0.036 & 0.071  & 0.140 & 0.290 \\
    Two-hot \cite{Newell_Hourglass_ECCV16}   & 0.009 & 0.018 & 0.037  & 0.090 & 0.271 \\
    Distribution-Aware \cite{Zhang_darkpose_CVPR20}  & \textbf{0.000} & 0.001 & 0.015  & 0.112 & 0.289 \\
\hline
    KeyPosS (Our) & \textbf{0.000} & \textbf{0.000} & \textbf{0.000}  & \textbf{0.000} & \textbf{0.000}\\ \hline
    \end{tabular}}
    \vspace{-0.15in}
    \label{table:upper}
\end{table}

\subsection{Theoretical Quantization Error}
\label{sec:keypoint_representation}
After a brief review of different encoding and decoding approaches, we explore various keypoint decoding strategies, including one-hot and two-hot heatmaps, the distribution-aware method, and our KeyPosS method (shown in Figure~\ref{fig:encoding}). Our goal is to analyze their technical characteristics and understand the underlying mechanisms that contribute to quantization error.

\noindent \textbf{(1) One-Hot Decoding.}~This technique is similar to traditional object detection methods and quantizes keypoint coordinates based on the heatmap resolution. The argmax operation selects the highest value as the keypoint location, which can result in notable errors due to misalignment between the ground truth location and the maximum response, especially when the downsampling ratio increases. The theoretical error can be estimated as half of the resolution's grid size.

\noindent \textbf{(2) Two-hot Decoding.}~Building upon the one-hot strategy, the two-hot approach includes both the maximum and second maximum responses in the prediction. This mitigates some of the localization error by providing a refined representation, but the quantization error may still be substantial. The incorporation of two maximum responses can offer a better approximation, but it does not fully overcome the discretization nature of the heatmap.

\noindent \textbf{(3) Distribution-Aware Decoding.}~The distribution-aware decoding strategy, proposed by Zhang et al. \cite{Zhang_darkpose_CVPR20}, reduces localization error by evaluating the Taylor series around the maximum activation position. This method considers the trends of neighboring points, providing an approximation that counters the discrete nature of heatmaps. The theoretical insight is that by taking into account the continuous distribution of the underlying data, the strategy becomes more robust with high-resolution heatmaps but suffers with low-resolution ones.

\noindent \noindent \textbf{(4) Our KeyPosS.}~Our proposed KeyPosS is a distance map-based heatmap that calculates the distance from a fixed anchor to the point of interest (POI). It employs True-range Multilateration to overcome the limitations of one-hot and two-hot techniques, resulting in more accurate localization. The method's accuracy is based on precise distance predictions, which produce dependable and accurate results across a variety of heatmap resolutions.

\begin{figure} [!ht]
\small
	\begin{center}
		\includegraphics[width=0.9\linewidth]{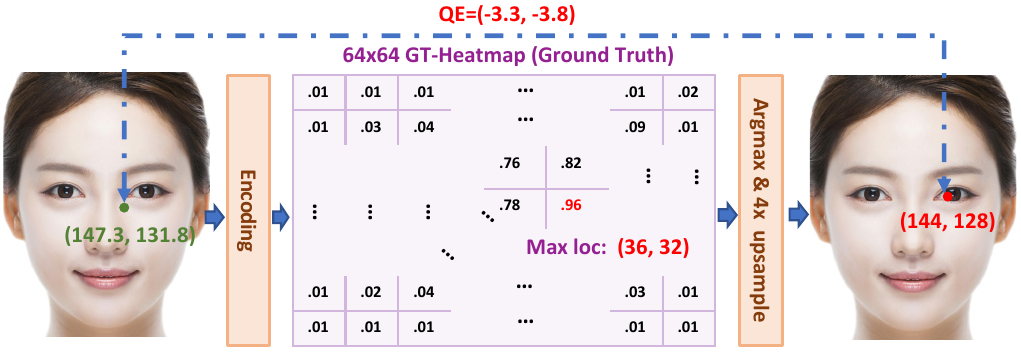}
	\end{center}
	\vspace{-0.15in}
	\caption{\small Unveiling the Theoretical Quantization Error Induced by the One-Hot (Argmax) Decoding Scheme. The term "QE" denotes the quantization error, and the "GT-heatmap" represents the heatmap encoded by the ground truth coordinate. This diagram illustrates how the upsampling step can amplify these errors, especially in smaller heatmap resolutions, and highlights that methods like two-hot and distribution-aware decoding provide alleviation but are not immune to significant performance degradation with low-resolution heatmaps.}
        \label{fig:introduction}
	\vspace{-0.15in}
 \label{fig:qeror-cause}
\end{figure}

\noindent \textbf{Insights into Quantization Error.}~Quantization error in these methods is fundamentally linked to the discretization of the continuous space and the argmax operation used in decoding. Figure \ref{fig:qeror-cause} illustrates that upsampling can exacerbate these errors. Strategies like two-hot \cite{Newell_Hourglass_ECCV16} and distribution-aware decoding \cite{Zhang_darkpose_CVPR20} alleviate errors, but low-resolution heatmaps can still cause significant performance degradation.

\begin{figure*} [!ht]
\small
	\begin{center}
		\includegraphics[width=0.9\linewidth]{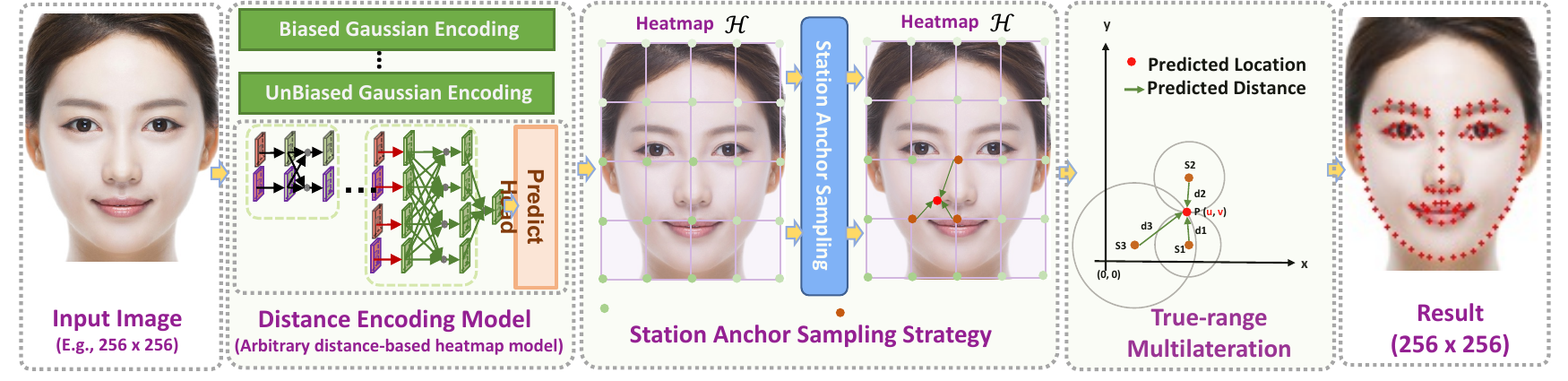}
	\end{center}
	\vspace{-0.1in}
	\caption{\small The pipeline of the proposed KeyPosS, consists of three components: (1) Distance Encoding Model, (2)  Station Anchor Sampling Strategy, and (3) True-range Multilateration. The KeyPosS scheme is versatile and can be adopted in \textit{any distance encoding-based approach.}}
	\label{fig:pipeline}
 \vspace{-0.1in}
\end{figure*}

\subsection{Experimental Quantization Error}
\label{sec:error_analysis}
After a rough intuitive theoretical analysis, we experimentally analyze quantization errors in multi-keypoint decoding strategies across different heatmap sizes. Table \ref{table:upper} details the experiment that uses GT keypoint coordinates as input. We evaluate the Normalized Mean Error (NME) between predicted and actual coordinates. Specifically, our analysis leverages the unbiased encoding method compatible with prevalent decoding techniques such as one-hot, two-hot, and distribution-aware.

\noindent \textbf{(1)~One-hot \& Two-hot Decoding.}~These decoding approaches demonstrate significant performance degradation as the heatmap scale decreases. While they may provide acceptable accuracy in high-resolution scenarios, the quantization error becomes more pronounced in low-resolution contexts, confirming the previous theoretical observations.

\noindent \textbf{(2)~Distribution-Aware Decoding.}~Acting as a compensation strategy, the distribution-aware decoding outperforms one-hot decoding at higher resolutions but experiences a sharp drop in performance under low-resolution conditions. This behavior emphasizes its ability to correct some errors in high-resolution settings, but also exposes its limitations when resolution is reduced.

\noindent \textbf{(3)~Our KeyPosS.}~Our proposed KeyPosS strategy, employing a multilateration method, showcases impressive robustness and adaptability across various heatmap scales. By interpreting the heatmap as a fusion of label and distance, it is able to shift seamlessly between high-resolution compensation-based methods and regression-based methods. This innovative approach transcends the boundaries of traditional strategies, suggesting that KeyPosS could provide a broader solution for different scenarios.

\noindent \textbf{Summary of Experimental Analysis.}~The experimental results elucidate the complex relationship between heatmap scale, decoding strategies, and resulting quantization errors. In general, the results of the experimental analysis resonate with the theoretical analysis, shedding light on the specific strengths and weaknesses of each method in real-world applications. Notably, the consistent performance of our KeyPosS across different scales highlights its potential as a groundbreaking approach in keypoint representation.

\subsection{Crafting KeyPosS Framework}
\label{sec:keyposs_framework}
Our KeyPosS approach, visualized in Figure \ref{fig:pipeline}, comprises several stages. Initially, an input image is processed through HRNet \cite{Sun_HRNet_CVPR19} to extract relevant features. Subsequently, a keypoint prediction head generates a distance map encoding the distance between the Point of Interest (POI) and station anchors. Formally, given an input image $I$, the backbone network $\mathcal{F}(\cdot)$ computes the feature map $f$, $f \in R^{HxWxC}$, using the network weights $W$. From this feature map, the keypoint prediction head constructs the final heatmap $\mathcal{H} \in R^{H\times W \times K}$.

\noindent \textbf{(1)~Distance Encoding Model.}~HRNet's high-resolution architecture \cite{Sun_HRNet_CVPR19} has been effective in dense label prediction tasks. However, to adapt to memory bandwidth constraints in mobile and edge devices, we propose a lightweight version of HRNet, which employs early fast downsampling. This downsampling uses convolutional layers with a fixed $3\times3$ kernel size. Table \ref{tab:fast-ds} shows the configurations for the number and stride of the convolutional layer.

\begin{figure} [!th]
\small
\begin{center}
\includegraphics[width=0.65\linewidth]{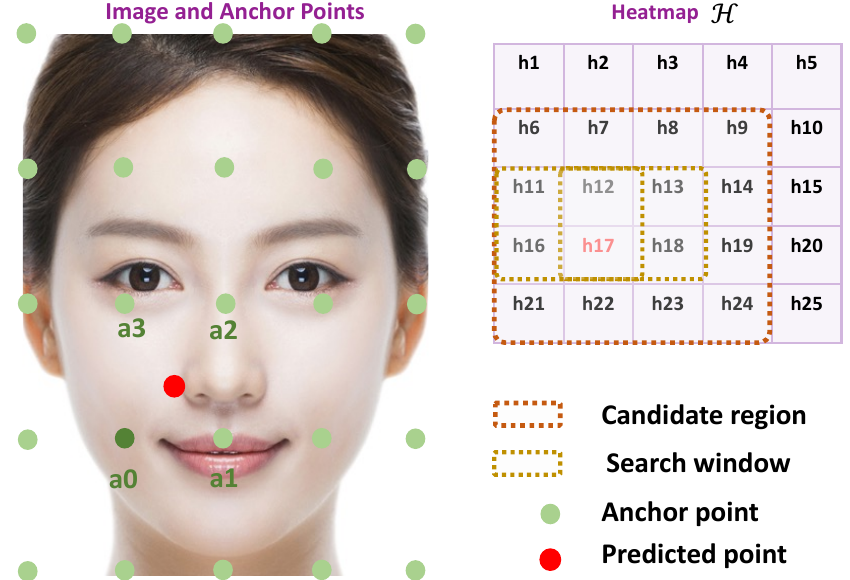}
\end{center}
\vspace{-0.1in}
\caption{\small Strategy of Anchor Sampling: A 5 $\times$ 5 heatmap illustrating the selection of the highest response points (e.g., h12, h13, h17, h18) within a search window centered around h17. The orange dashed box represents the search window, which is slided with a step size of 1, ensuring non-collinearity and optimizing the Normalized Mean Error (NME) using 3 or 4 anchor points.}
\label{fig:anchorsampling}
\vspace{-0.2in}
\end{figure}

\noindent \textbf{(2)~Station Anchor Sampling Strategy.}~The accuracy of the POI location prediction hinges on an appropriate selection of station anchors. These are chosen based on two criteria: a) selection of over three non-collinear anchors and b) prediction accuracy decreases as distance increases.

To meet these requirements, we propose a filtering-based anchor sampling strategy. For a given heatmap $\mathcal{H}$, we apply a fixed filter kernel $K$ to the original heatmap using the operation $\otimes$, which is performed with a stride of 1 and a kernel value of 1. We then find the region in the input-filtered heatmap with the highest value, which is used in the True-range Multilateration algorithm. By adjusting the kernel size, we can control the number of sampled anchors, ensuring their non-collinearity.

As shown in Figure \ref{fig:anchorsampling}, consider a 5x5 heatmap, the process begins by finding the point on the heatmap with the highest value, such as point h17. A search window of range [-2, +2] is centered around h17 (shown by the orange dashed box). This window is slided with a step size of 1 to find the four points with the highest responses, which might be points h12, h13, h17, and h18, as anchor points for decoding. This method ensures non-collinearity of the selected anchor points. Experimental results show that the Normalized Mean Error (NME) is minimized with 3 or 4 anchor points.

\noindent \textbf{(3)~True-range Multilateration.}~The True-range Multilateration algorithm is widely employed in various positioning applications. Given a set of station anchors $\mathcal{A}=\{A_{(x_{0},y_{0})}, A_{(x_{1},y_{1})}, \dots, A_{(x_{n},y_{n})}\}$ and their corresponding distances $\mathcal{D}=\{d_{0}, d_{1},\dots d_{n}\}$ to the POI $\boldsymbol{p}$, a system of non-linear equations is formulated as follows:
\begin{equation}
\begin{cases}
(x_{0} - u)^2 + (y_{0} - v)^2= d_{0}^2 \\
(x_{1} - u)^2 + (y_{1} - v)^2= d_{1}^2 \\
\vdots \\
(x_{n} - u)^2 + (y_{n} - v)^2= d_{n}^2 
\end{cases}
\end{equation}
where $x_{n}$, $y_{n}$, and $d_{n}$ denote the $x$, $y$ coordinates of the station anchor $A_{(x_{n},y_{n})}$ and the distance from the station anchor to the POI $\boldsymbol{p}$. The system of non-linear equations can be solved using the least square method, which is transformed into the form:
\begin{equation}
\boldsymbol{Y}=\hat{\boldsymbol{\beta}} \boldsymbol{X}
\end{equation}
In this equation, $\hat{\boldsymbol{\beta}}$ represents the POI location to be estimated, and $\boldsymbol{X}$, $\boldsymbol{Y}$, and $\hat{\boldsymbol{\beta}}$ are defined as follows:
\begin{equation}
\boldsymbol{X}=
\left[
\begin{array}
{cc}
x_{n} - x_{1} & y_{n} - y_{1}\\
x_{n} - x_{2} & y_{n} - y_{2}\\
\vdots \\
x_{n} - x_{n-1} &(y_{n} - y_{n-1}) 
\end{array}
\right]
\end{equation}
\begin{equation}
\boldsymbol{Y}=
\left[
\begin{array}
{c}
(d_{1}^2 - d_{n}^2+x_{n}^2+y_{n}^2-x_{1}^2-y_{1}^2)/2\\
(d_{2}^2 - d_{n}^2+x_{n}^2+y_{n}^2-x_{2}^2-y_{2}^2)/2 \\
\vdots \\
(d_{n-1}^2 - d_{n}^2+x_{n}^2+y_{n}^2-x_{n-1}^2-y_{n-1}^2)/2 
\end{array}
\right]
\end{equation}
\begin{equation}
\hat{\boldsymbol{\beta}} =
\left[
\begin{array}
{c}
u\\v
\end{array} 
\right]
\end{equation}
The predicted location $\hat{\boldsymbol{\beta}}$ can be obtained by:
\begin{equation}\label{equ:solved}
\hat{\boldsymbol{\beta}} = (\boldsymbol{X}^{T}\boldsymbol{X})^{-1}\boldsymbol{X}^{T}\boldsymbol{Y}
\end{equation}

\begin{table}[!t]
\small
\centering
 \caption{\small Setup of convolutional layers for expedited downsampling. The symbol `-' denotes the absence of a specific convolutional layer.}
 \vspace{-0.1in}
\begin{tabular}{c|ccccc}
 \hline
 Resolution & \multicolumn{1}{c|}{64}   & \multicolumn{1}{c|}{32}   & \multicolumn{1}{c|}{16}   & \multicolumn{1}{c|}{8}    & 4    \\ \hline

Conv1.Stride     & \multicolumn{1}{c|}{2} & \multicolumn{1}{c|}{4} & \multicolumn{1}{c|}{4}  & \multicolumn{1}{c|}{4} & 4 \\ \hline
Conv2.Stride     & \multicolumn{1}{c|}{2} & \multicolumn{1}{c|}{2} & \multicolumn{1}{c|}{4} & \multicolumn{1}{c|}{4} & 4 \\ \hline

Conv3.Stride     & \multicolumn{1}{c|}{-} & \multicolumn{1}{c|}{-} & \multicolumn{1}{c|}{-} & \multicolumn{1}{c|}{2} & 2 \\ \hline

Conv4.Stride     & \multicolumn{1}{c|}{-} & \multicolumn{1}{c|}{-} & \multicolumn{1}{c|}{-} & \multicolumn{1}{c|}{-} & 2 \\ \hline
\end{tabular}
\label{tab:fast-ds}
\vspace{-0.2in}
\end{table}

\begin{table*}[!th] 
\small
\centering
\caption{\small Comparison with the State-of-the-Art methods. The results are in NME (\%). The best results are highlighted with bold text font.}
\vspace{-0.1in}
\begin{tabular}{c|c|c|c|c|c|c|c}
\hline
Method & Year  & Pretrained & WFLW \cite{Wu_LaB_CVPR18}   & AFLW \cite{Kostinger_AFLW_ICCVW11} & 300W \cite{Sagonas_300W_IVC16} & Parameters & GFlops\\ \hline 
DAC-CSR \cite{Feng_DACCSR_CVPR17}      & CVPR2017           & -    & -         & 2.27 & -      &  -  &-\\ 
TSR \cite{Lv_TSR_CVPR17}               & CVPR2017           & -    & -         & 2.17 & 4.99    & -  &-\\ 
LAB \cite{Wu_LaB_CVPR18}               & CVPR2018           & -    & 5.27      & 1.85  & 3.49   &  24.1M+28.3M  & 26.7+2.4\\ 
Wing  \cite{Feng_Wingloss_CVPR18}      & CVPR2018           & Y    & 4.99      & 1.47  & -      &  91.0M        & 5.5\\ 
ODN \cite{Zhu_OAD_CVPR19}              & CVPR2019           & Y    & -         & 1.63  & 4.17   &  -            & -   \\ 
DeCaFa \cite{Dapogny_DeCaFA_ICCV19}    & ICCV2019           & -    & 4.62      & -     & 3.39   &  10M            & -   \\ 
DAG \cite{Li_DAG_ECCV20}               & ECCV2020           & Y    & 4.21      & -     & 3.04   &  -            & -    \\ \hline

AWing \cite{Wang_AWingLoss_ICCV19}     & ICCV2019           & N    & 4.36      & 1.53  & 3.07   &  24.1M        & 26.7    \\ 
AVS \cite{Qian_AVS_ICCV19}             & ICCV2019           & N    & 4.39      & -     & 3.86   &  28.3M        & 2.4   \\ 
ADA \cite{Chandran_attention_CVPR2020} & CVPR2020           & -    & -         & -     & 3.50   &  -            & -   \\ 
LUVLi \cite{Kumar_LUVLi_CVPR20}        & CVPR2020           & N    & 4.37      & 1.39  & 3.23   &  -            & -   \\ 
PIPNet-18 \cite{Jin_PIPNet_IJCV21}     & IJCV2021           & Y    & 4.57      & 1.48  & 3.36   & 12.0M         & 2.4    \\ 
PIPNet-101 \cite{Jin_PIPNet_IJCV21}    & IJCV2021           & Y    & 4.31      & 1.42  & 3.19   & 45.7M         & 10.5   \\ 
DTLD \cite{Li_transformer_CVPR22}      & CVPR2022          & Y    & 4.08       & 1.38  & \textbf{2.96}    &  -            & -      \\ 
RePFormer \cite{li2022repformer}                      & IJCAI2022                   & Y    & 4.11    & 1.43  & 3.01  & -         & -\\ 
SLPT \cite{xia2022sparse}                           & CVPR2022                   & -    & 4.14    & -  & 3.17  & 9.98M         & -\\ 
EF-3-ACR \cite{fard2022acr}                       & ICPR2022                   & -    & -    & -  & 3.75  & -         & -\\ 
ADNet-FE5 \cite{huang2023freeenricher}                      & AAAI2022                   & Y    & 4.1    & -  & 2.87  & -         & -\\ 
ResNet50-FE5 \cite{huang2023freeenricher}                   & AAAI2022                   & -    & -    & -  & 4.39  & -         & -\\ 
HRNet-FE5 \cite{huang2023freeenricher}                      & AAAI2022                   & -    & -    & -  & 3.46  & -         & -\\ 
STAR Loss \cite{zhou2023star}                      & CVPR2023                   & -    & 4.02    & -  & 2.87  & 13.37M         & -\\ 
RHT-R \cite{wan2023precise}                          & TIP2023                    & -    & 4.01    & 1.99  & 3.46  & -         & -\\ \hline
KeyPosS (Biased)                       & 2023                   & Y    & \textbf{4.0}    & \textbf{1.35}  & 3.39  & 9.7M         & 4.7\\ 
KeyPosS (Unbiased)                     & 2023                  & Y    & \textbf{4.0}    & \textbf{1.35}  & 3.34   & 9.7M         & 4.7\\ \hline
\end{tabular}
\label{tab:SOTA}
\end{table*}

\noindent \textbf{(4)~Training Details.}~KeyPosS utilizes Mean Square Error (MSE) loss \cite{Feng_Wingloss_CVPR18} for supervising model training. The MSE loss is calculated as follows:
\begin{equation}
\mathcal{L} = \sum_{i=1}^{K}||\beta_{i} - \hat{\beta}{i}||_{2}
\end{equation}
where $\beta_{i}$ and $\hat{\beta}{i}$ represent the ground truth and predicted locations, respectively, with $\hat{\beta}{i}$ computed using Equation \ref{equ:solved}, and $K$ representing the total facial landmarks. The model, during training, uses face images and Ground Truth (GT) heatmaps - encoded from GT coordinates - as input and produces a set of heatmaps. Although HRNet is used as the backbone of our experiments, other suitable models can be used as replacements.
\section{Experiments}
We first present the dataset and evaluation metrics (Sec.\ref{sec:dataset}), followed by detailing the implementation of our KeyPosS (Sec.\ref{sec:implement}). Subsequently, we compare our approach with state-of-the-art methods (Sec.\ref{sec:ext-sota}) and previous decoding techniques (Sec.\ref{sec:ext-previous}). Finally, we offer ablation studies (Sec.\ref{sec:ablation}), visual results (Sec.\ref{sec:ext:visul}), and an efficiency analysis (Sec.~\ref{sec:ext-effient}).

\vspace{-2mm}
\subsection{Dataset and Metric}
\label{sec:dataset}
\noindent \textbf{WFLW \cite{Wu_LaB_CVPR18}}. The Wider Facial Landmarks in-the-wild (WFLW) dataset includes 10,000 face images (7,500 for training and 2,500 for testing), each annotated with 98 landmarks. With substantial pose, expression, and occlusion variations, the test set is further divided into six evaluation categories: pose, expression, illumination, make-up, occlusion, and blur.

\noindent \textbf{AFLW \cite{Kostinger_AFLW_ICCVW11}}. The Annotated Facial Landmarks in the Wild (AFLW) dataset provides a large-scale collection of annotated face images from Flickr. These images display extensive variation in appearance (e.g., pose, expression, ethnicity, age, gender) and general imaging and environmental conditions. Overall, approximately 25k faces are annotated, with up to 21 landmarks per image.

\noindent \textbf{300W \cite{Sagonas_300W_IVC16}}. The 300W dataset comprises 300 indoor and 300 outdoor in-the-wild images, covering a wide range of identities, expressions, illumination conditions, poses, occlusion, and face sizes. Notably, each face in the dataset is annotated with 68 landmarks.

\noindent \textbf{COCO-WholeBody \cite{Jin_COCOWholebody_ECCV20}}. The COCO-WholeBody dataset extends the COCO dataset with whole-body annotations. It features four types of bounding boxes (person, face, left-hand, right-hand) and 133 keypoints (17 body, 6 feet, 68 face, 42 hands) per person in an image. In this study, we only focus on face landmarks annotation for performance evaluation.

\noindent \textbf{Evaluation Metric}. For consistency with previous works, we adopt the Normalized Mean Error (NME) metric to evaluate our model. Typically, it is calculated as follows:
\vspace{-0.1in}
\begin{equation}
\mathcal{L} = \sum_{i=1}^{N}||\beta_{i} - \hat{\beta}_{i}||_{2}
\end{equation}
where $\beta$ and $\hat{\beta}$ represent the ground truth and predicted results, respectively. $D$ denotes the normalized distance, and $N$ is the number of landmarks.

\subsection{Implementation Details}
\label{sec:implement}
Our implementation of KeyPosS is built on the MMpose key point detection framework\footnote{https://github.com/open-mmlab/mmpose}.
We preprocess all datasets by cropping the facial region according to the provided bounding box and resizing it to $256 \times 256$ resolution.
Data augmentation techniques including random translation, horizontal flipping, rotation, and scaling are applied to improve model robustness and generalization.
We experiment with varying heatmap sizes of $64\times64$, $32\times32$, $16\times16$, $8\times8$, and $4\times4$ pixels, which allowed us to understand the tradeoffs between resolution and computational requirements.

Specifically, the end-to-end KeyPosS framework uses the Adam optimizer \cite{Kingma_Adam_ICLR15} for training.
The backbone is initialized with an HRNet \cite{Sun_HRNet_CVPR19} pre-trained on ImageNet \cite{ILSVRC15}.
The training employs the Adam optimizer with a linear-step learning rate decay schedule.
The initial learning rate is set to 2e-3 and reduced to 1e-5 gradually by the 100th epoch. 
We use a batch size of 64, adjusting to $64 \times 64$ for the extended 1000 epoch training.
The Normalized Mean Error (NME) is used as the evaluation metric.
All experiments are performed on a 4 $\times$ NVIDIA V100 GPU server.

\subsection{Comparison with the State-of-the-Art}
\label{sec:ext-sota}
Table \ref{tab:SOTA} compares KeyPosS against state-of-the-art methods on four key datasets. On AFLW and WFLW datasets, KeyPosS stands out, outperforming one-hot and two-hot heatmap-based baselines by considering neighboring responses to correct the location of the maximum response. Meanwhile, coordinate regression-based baselines fall behind as they often lose spatial details during forward passes, hindering high-precision localization. It is noted that KeyPosS's amalgamation of classification and regression excels here, especially in high-resolution settings.

On the more complex 300W dataset, KeyPosS achieves 3.39\% NME, highlighting two challenges: 1) the complexity of the 300W dataset, and 2) underfitting due to limited training data. While acknowledging DTLD's impressive performance, KeyPosS's competitive results on WFLW and AFLW with fewer parameters and higher computational efficiency should also be noted.

Additionally, the results on the COCO-WholeBody dataset, presented in Table \ref{table:ms-heatmap}, further demonstrate KeyPosS's capabilities. Since recent works lack experiments on this dataset, we provide a direct comparison in our ablation studies.

\begin{figure} [!ht]
\small
\begin{center}
\vspace{-0.05in}
    \includegraphics[width=0.85\linewidth]{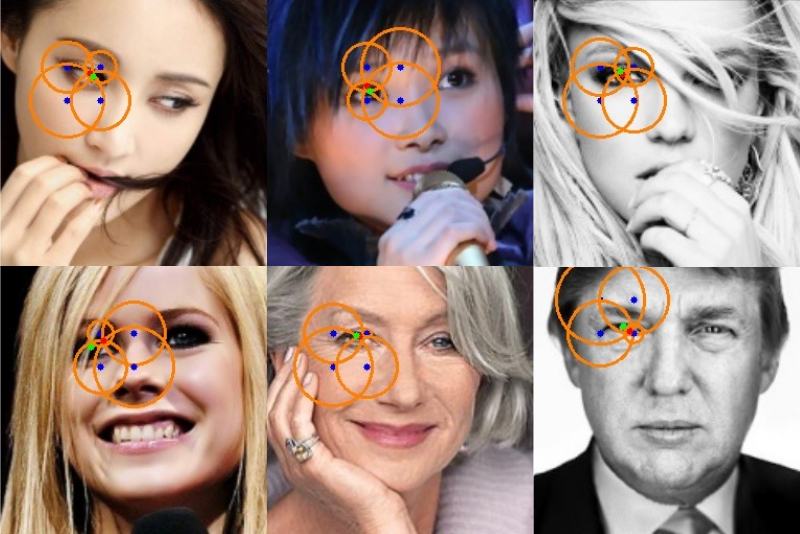}
\end{center}
\vspace{-0.15in}
\caption{\small Visualization of True-range Multilateration. The \textcolor{blue}{blue}, \textcolor{green}{green}, and \textcolor{red}{red} dots indicate the anchor station, actual position, and predicted location respectively. Predicted distances are represented by \textcolor{orange}{orange circles}.}
\vspace{-0.15in}
\label{fig:tril-vis}
\end{figure}

\begin{table}[!ht] 
\small
\centering
\caption{\small Impact of Anchor Sampling on Model Performance.}
\vspace{-0.15in}
\begin{tabular}{c|c|c|c|c|c}
\hline
Kernel size & $2\times2$   & $3\times3$   & $4\times4$   & $5\times5$  & $6\times6$  \\ \hline
$64^2$          & 4.00  & 4.00  & \textbf{3.98}  & 3.99 & 3.99 \\ \hline
$32^2$          & 4.30  & 4.29  & \textbf{4.27}  & 4.28 & 4.30 \\ \hline
$16^2$          & \textbf{4.94}  & 4.94  & 4.95  & 4.96 & 5.00 \\ \hline
$8^2$           & \textbf{7.67}  & 7.69  & 7.72  & 7.80 & 9.41 \\ \hline
$4^2$           & \textbf{12.04} & 26.57 & 18.74 & -    & -    \\ \hline
\end{tabular}
\label{tab:sampling}
\vspace{-0.10in}
\end{table}

\begin{table*}
\small
    \centering
    \caption{\small Performance of keypoint representations at different scales of the heatmap.}
    \vspace{-0.1in}
    \setlength{\tabcolsep}{1.35mm}{
    \begin{tabular}{l|c|c|c|c|c|c|c|c|c|c|c|c|c|c|c}
    \hline
    Dataset               & \multicolumn{5}{|c|}{WLFW~\cite{Wu_LaB_CVPR18}}  & \multicolumn{5}{|c|}{AFLW~\cite{Kostinger_AFLW_ICCVW11}}  & \multicolumn{5}{|c}{COCO-WholeBody~\cite{Jin_COCOWholebody_ECCV20}} \\ \hline
    Resolution            & $64^{2}$  & $32^{2}$ & $16^{2}$ & $8^{2}$   & $4^{2}$ & $64^{2}$  & $32^{2}$ & $16^{2}$ & $8^{2}$   & $4^{2}$ & $64^{2}$  & $32^{2}$ & $16^{2}$ & $8^{2}$   & $4^{2}$\\ \hline
    One-hot                              & 4.28 & 5.23 & 7.59  & 14.74 & 28.8 & 1.57 & 2.11 & 3.57  & 6.52 & 13.06 & 5.96 & 7.14 & 10.50  & 17.98 & 37.7 \\
    Two-hot \cite{Newell_Hourglass_ECCV16}      & 4.06 & 4.55 & 5.83  & 12.3 & 27.4 & 1.41 & 1.66 & 2.54  & 5.08 & 12.77 & 5.59 & 6.08 & 8.11  & 14.28 & 34.54 \\
     KeyPosS (Biased)   & \textbf{3.99} & \textbf{4.30} & \textbf{5.03}  & \textbf{9.96}  & \textbf{19.48} & \textbf{1.35} & \textbf{1.49}  & \textbf{2.12}   & \textbf{4.18} & \textbf{10.04} & \textbf{5.45}  & \textbf{5.64}& \textbf{6.93}  & \textbf{10.98} & \textbf{27.85}\\ 
    \hline\hline
    Distribution-Aware \cite{Zhang_darkpose_CVPR20}  & \textbf{3.96} & \textbf{4.25} & 5.38  & 11.83 & 26.05 & 1.35 & 1.43 & 1.77  & 4.83 & 13.28 & \textbf{5.10} & \textbf{5.71} & {6.58}  & 15.71 & 40.12 \\
    KeyPosS (Unbiased)       & 3.98 & 4.27 & \textbf{4.94}  & \textbf{7.68} & \textbf{12.04} & \textbf{1.35} & \textbf{1.43} & \textbf{1.73}  & \textbf{3.34} & \textbf{4.54} & {5.17} & {5.80} & \textbf{6.19}  & \textbf{10.37} & \textbf{27.59}\\ \hline
    \end{tabular}}
    \label{table:ms-heatmap}
\end{table*}

\begin{figure*} [!ht]
\centering
\small
\begin{center}
    \includegraphics[width=0.95\linewidth]{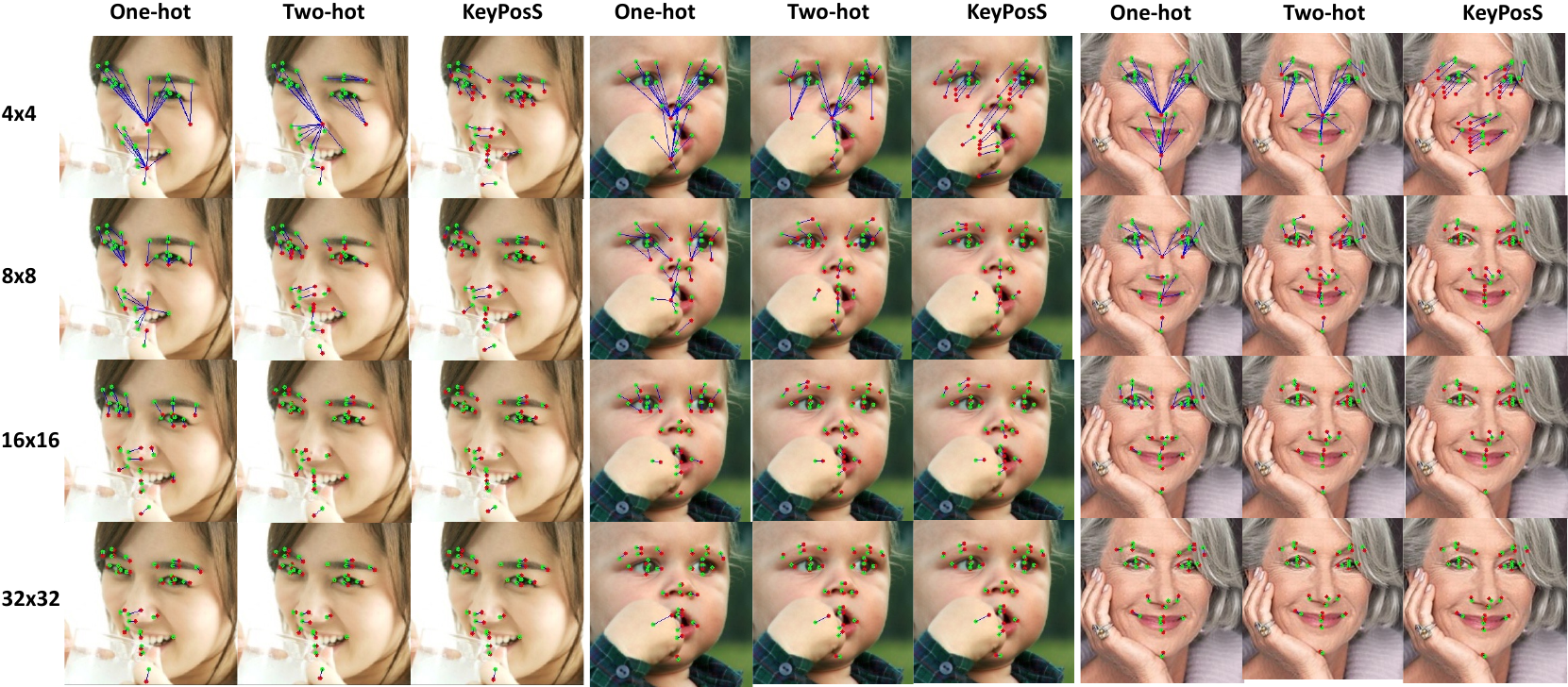}
\end{center}
\vspace{-0.1in}
\caption{\small Depiction of heatmap size's influence on prediction outcomes. Results are shown at varying heatmap scales, in descending order. One-hot, two-hot, and KeyPosS prediction results are grouped separately. Ground truth is represented by \textcolor{green}{green dots}, and predicted results by \textcolor{red}{red dots}. Discrepancies between predictions and ground truth are highlighted with \textcolor{blue}{blue lines}.}
\label{fig:vis-landmarks}
\vspace{-0.05in}
\end{figure*}

\vspace{-2mm}
\subsection{Comparison with Decoding Schemes}
\label{sec:ext-previous}
KeyPosS is a plug-in decoding scheme that can be readily incorporated into existing encoding methods. To validate its effectiveness and robustness, we adopt two classic biased and unbiased encoding strategies from DarkPose \cite{Zhang_darkpose_CVPR20}, benchmarking against one-hot, two-hot, and distribution-aware decoding baselines. The upper part of Table \ref{table:ms-heatmap} shows results for \textit{Biased Encoding}, following the original one-hot/two-hot methods, while the bottom part shows results for \textit{Unbiased Encoding}.

With biased encoding, KeyPosS outperforms one-hot and two-hot methods across all resolutions on COCO-WholeBody, WFLW, and AFLW. Notably, KeyPosS achieves significant improvements for resolutions from $16\times16$ to $4\times4$, demonstrating its robustness. Although the two-hot baseline yields reasonable performance for high-resolution heatmaps, it still lags behind KeyPosS, showing KeyPosS's superiority over this relatively simple approach.

The results with unbiased encoding largely mirror those of biased encoding. However, KeyPosS holds its own against distribution-aware methods at $64\times64$ and $32\times32$ resolutions. This reveals some limitations of KeyPosS, such as its heavy dependence on distance prediction accuracy. The true-range multilateration problem cannot be precisely solved with least squares, and the predicted heatmap can contain multiple peaks, leading to incorrect solutions - areas for future work.

\vspace{-2mm}
\subsection{Ablation Studies}
\label{sec:ablation}
\noindent \textbf{Effect of Anchor Sampling}.
To understand the impact of the stationary anchor sampling strategy, we performed ablation studies focusing on kernel size and heatmap resolution. The results, as illustrated in Table~\ref{tab:sampling}, confirm that a small $2 \times 2$ kernel size delivers optimal performance across varying resolutions. This discovery underscores the significance of focusing on the highest response region for precise localization. Interestingly, it also suggests that simply increasing the number of stationary anchors does not lead to corresponding enhancements in performance.

\noindent \textbf{Exploration of True-range Multilateration}.
A comprehensive analysis of True-range Multilateration's role in localization was conducted by visualizing the distance predictions and their intersecting regions. As shown in Table~\ref{tab:sampling}, the investigation revealed that location errors are primarily attributed to inaccuracies in distance prediction. Generally, a shorter span from the stationary anchor results in more precise distance approximations. However, as illustrated in Figure \ref{fig:tril-vis}, even the proximity of the closest anchor can introduce substantial errors. This highlights that stationary anchor sampling, while critical to our method, still poses significant challenges and offers opportunities for further refinement.

\vspace{-2mm}
\subsection{Visualization Analysis}
\label{sec:ext:visul}
To comprehensively study KeyPosS's adaptability, particularly in real-world applications, we undertook a visualization analysis comparing it against one-hot and two-hot baselines. The experiments included comparative assessments with one-hot and two-hot baselines. Since the existing methods already demonstrate acceptable results in high-resolution heatmap configurations, our focus shifted to the more challenging low-resolution settings. These comparisons are visually depicted in Figure \ref{fig:vis-landmarks}.

Upon examination, it is clear that the traditional one-hot and two-hot baselines falter at lower resolutions, such as $4\times4$ and $8\times8$, failing to provide accurate predictions. In contrast, our proposed KeyPosS maintains functionality even at the $4\times4$ resolution and excels at $8\times8$ and higher resolutions. A significant observation is that when the resolution exceeds $32\times32$, the predicted results across various methods converge, becoming virtually indistinguishable. This finding underscores the impressive capabilities of KeyPosS, proving its effectiveness not only at low resolutions but also in scenarios requiring standard precision. This visualization analysis further accentuates the innovative nature of KeyPosS and its substantial potential in a wide array of applications.

\subsection{Efficiency at Different Heatmap Scales}
\label{sec:ext-effient}
As shown in Table~\ref{tab:efficiency}, KeyPosS has demonstrated robust performance across various heatmap resolutions, highlighting its adaptability and efficiency. Through comprehensive analysis across scales, we found that downsampling the feature maps before HRNet's first stage leads to an exponential reduction in FLOPs without compromising accuracy. Remarkably, KeyPosS maintains strong capabilities even at low resolutions of $16\times16$ and $8\times8$. In practical use, these correspond to a mere 0.3 and 0.07 GFLOPs respectively. Such low computational demands showcase KeyPosS's potential as an exceptionally lightweight facial landmark detection solution, making it valuable for real-world applications with constrained computational resources.

\begin{table}[]
\small
\centering
\caption{\small Efficiency at Different Scale.}
\vspace{-0.1in}
\begin{tabular}{c|ccccc}
\hline
input size   & \multicolumn{5}{c}{$256\times256$}                                                                                   \\ \hline
heatmap size  & \multicolumn{1}{c|}{$64^2$}   & \multicolumn{1}{c|}{$32^2$}   & \multicolumn{1}{c|}{$16^2$}   & \multicolumn{1}{c|}{$8^2$}    & {$4^2$}    \\ \hline

Flops(GFLOPs) & \multicolumn{1}{c|}{4.75} & \multicolumn{1}{c|}{1.19} & \multicolumn{1}{c|}{0.3}  & \multicolumn{1}{c|}{0.07} & 0.03 \\ \hline
Params(M)     & \multicolumn{1}{c|}{9.74} & \multicolumn{1}{c|}{9.74} & \multicolumn{1}{c|}{9.74} & \multicolumn{1}{c|}{9.66} & 9.64 \\ \hline
\end{tabular}
\label{tab:efficiency}
\vspace{-0.15in}
\end{table}

\section{Conclusion}
We have introduced KeyPosS, an innovative facial landmark detection framework that incorporates True-range Multilateration, commonly used in GPS systems, to accurately localize 2D image Points of Interest (POIs). Through a fully convolutional network, KeyPosS generates a distance map from POIs to designated anchors and triangulates precise POI coordinates. Benchmarked against state-of-the-art models across diverse datasets, KeyPosS has demonstrated superior performance and efficiency, positioning it as a breakthrough approach for facial landmark detection, especially in resource-constrained and low-resolution scenarios. Moving forward, we aim to expand KeyPosS's capabilities to object detection and pose estimation tasks. The KeyPosS code is publicly available at \hyperlink{blue}{https://github.com/zhiqic/KeyPosS}.

\section*{Acknowledgments}
The contributions of Zhi-Qi Cheng in this project were supported by the Army Research Laboratory (W911NF-17-5-0003), the Air Force Research Laboratory (FA8750-19-2-0200), the U.S. Department of Commerce, National Institute of Standards and Technology (60NANB17D156), the Intelligence Advanced Research Projects Activity (D17PC00340), and the US Department of Transportation (69A3551747111).  Intel and IBM Fellowships also provided additional support for Zhi-Qi Cheng's research work. The views and conclusions contained herein represent those of the authors and not necessarily the official policies or endorsements of the supporting agencies or the U.S. Government.

\bibliographystyle{ACM-Reference-Format}
\bibliography{egbib}
\end{document}